\documentclass[conference]{IEEEtran}
\usepackage{times}

\IEEEoverridecommandlockouts    % enables disabled commands like \thanks

\usepackage[numbers]{natbib}
\usepackage{multicol}
\usepackage[bookmarks=true]{hyperref}

\usepackage{amsmath}
\usepackage{amsthm}
\usepackage{amssymb}
\usepackage{graphicx}
\usepackage[table]{xcolor}
\usepackage{multirow}
\usepackage{color}  % For Highlighting
\usepackage[normalem]{ulem}
\usepackage{soul}
\usepackage[ruled]{algorithm2e}
\usepackage{boldline}
\usepackage{hhline}

\newcounter{RamCount}
\newcounter{DavidCount}
\newcounter{BrentCount}

\newcommand{\Real}{\mathbb{R}}
\newcommand{\koop}{\mathcal{K}}

\usepackage{outlines}

\begin{document}
\setlength{\textfloatsep}{8pt}% decrease vertical space after figures

\title{Koopman-based Control of a Soft Continuum Manipulator Under Variable Loading Conditions}

% You will get a Paper-ID when submitting a pdf file to the conference system
% \author{Author Names Omitted for Anonymous Review. Paper-ID [1326]}
\author{Daniel~Bruder,~     % REMOVE IEEE AFFILIATIONS FOR ARXIV VERSION
        Xun~Fu,~
        R.~Brent~Gillespie,~
        C.~David~Remy,~
        and~Ram~Vasudevan% <-this % stops a space
\thanks{D. Bruder, X. Fu, R.B. Gillespie, and R. Vasudevan are with the Department
of Mechanical Engineering, University of Michigan, Ann Arbor,
MI, 48109 USA 
(e-mail: bruderd@umich.edu, xunfu@umich.edu, brentg@umich.edu, ramv@umich.edu).}% <-this % stops a space
\thanks{C.D. Remy is with the Institute for Nonlinear Mechanics, University of Stuttgart, Stuttgart, Germany (e-mail: david.remy@inm.uni-stuttgart.de).}% <-this % stops a space
\thanks{{A supplementary video can be found here: \href{https://youtu.be/g2yRUoPK40c}{https://youtu.be/g2yRUoPK40c}}}
}

\maketitle

\begin{abstract}
Controlling soft continuum manipulator arms is difficult due to their infinite degrees of freedom, nonlinear material properties, and large deflections under loading.
This paper presents a data-driven approach to identifying soft manipulator models that enables consistent control under variable loading conditions.
This is achieved by incorporating loads into a linear Koopman operator model as states and estimating their values online via an observer within the control loop.
Using this approach, real-time, fully autonomous control of a pneumatically actuated soft continuum manipulator is achieved.
In several trajectory following experiments, this controller is shown to be more accurate and precise than controllers based on models that are unable to explicitly account for loading.
The manipulator also successfully performs pick and place of objects with unknown mass, demonstrating the efficacy of this approach in executing real-world manipulation tasks.
\end{abstract}

\IEEEpeerreviewmaketitle

% main body
\section{Introduction}  \label{sec:intro}

Soft continuum manipulator arms are lightweight, cheap to make, and their inherent compliance carries the promise to enable safe interaction with humans, delicate objects, and the environment ~\cite{immega1995ksi,hannan2001elephant,mcmahan2005design,grissom2006design,cianchetti2013stiff,mahl2014variable,hughes2016soft,phillips2018dexterous}.
These properties would make them ideal platforms for tasks involving physical human-robot interaction such as feeding~\cite{miller1998assistive} or handling products in a warehouse~\cite{correll2016analysis}.
Yet, so far, the real-world application of soft manipulators has been limited.
This is due to the difficulty involved in controlling such systems, as they exhibit infinite degrees-of-freedom, nonlinear material properties, and large deflections under loading \cite{george2018control}.
These characteristics greatly complicate manipulation tasks such as pick and place, which require consistent control performance regardless of payload.
%As a result, teleoperation remains the dominant control strategy for performing tasks such as object manipulation.
Control-oriented models that describe soft manipulator behavior under varying loading conditions could enable the automation of such tasks.
% To automate such tasks, control-oriented models are needed that are capable of describing soft manipulator behavior under various loading conditions.%, not just when unloaded.

%% Overview block diagram figure
\begin{figure}
    \centering
    \includegraphics[width=\linewidth]{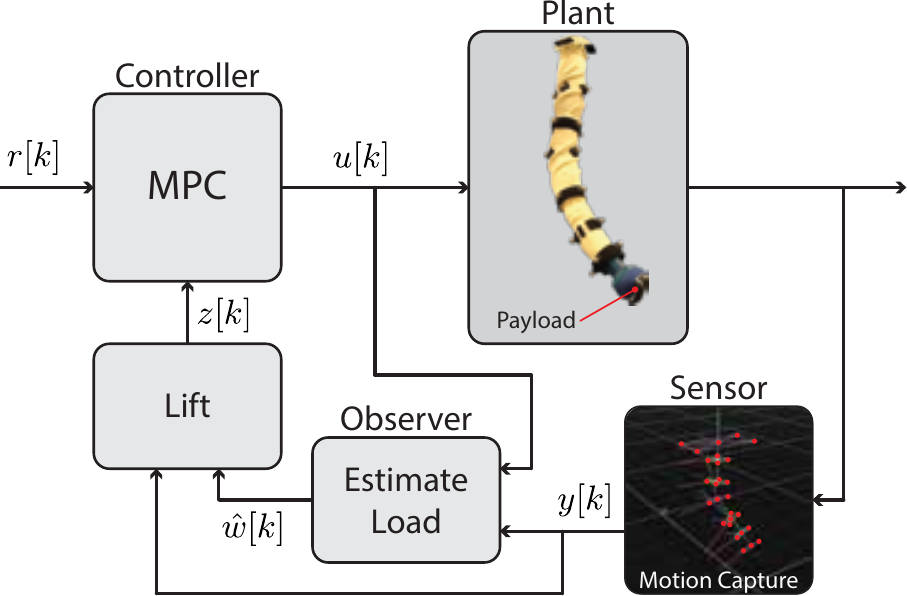}
    \caption{
    % Closed-loop control of a soft continuum manipulator is achieved by incorporating the payload into a linear Koopman model and estimating its value online. 
    A soft continuum manipulator is tasked with following a reference trajectory $r[k]$ while carrying an unknown payload. At each time step a model predictive controller computes the optimal input $u[k]$ to follow the reference trajectory based on a linear Koopman operator model, and the position of the end effector $y[k]$ is measured by a motion capture system. An observer computes a payload estimate $\hat{w}$ based on the previously measured inputs and outputs, which is then incorporated into the state of the model $z[k]$ via a lifting function.}
    \label{fig:overview}
\end{figure}

%% Justification for data-driven modeling
Recently, data-driven modeling techniques have emerged as a powerful tool to address the challenge of modeling soft continuum manipulators.
A primary benefit of these techniques is that 
% unlike physics-based modeling, 
a description of an input/output relationship can be obtained from system observations without explicitly defining a system state.
This is especially useful for obtaining reduced-order models of continuum robots that have essentially infinite-dimensional kinematics, without making simplifying physical assumptions such as 
piecewise constant curvature \cite{webster2010design}, pseudo-rigid-body mechanics \cite{howell1996evaluation}, quasi-static behavior \cite{bruder2018iros, thuruthel2018model, gravagne2003large, trivedi2008geometrically}, or simplified geometry \cite{sedal2019comparison, bruder2017model, sedal2017constitutive, bishop2012parallel}.
A potential downside of data-driven modeling is that it requires system behavior to be observed under a wide range of operating conditions, including those that may be dangerous to a robot or its surroundings.
Fortunately, compared to conventional rigid-bodied manipulators, soft manipulators pose much less of a physical threat to themselves and their surroundings.
It is hence possible to automatically and safely collect data under a wide range of operating conditions,
making soft robots well suited for data-driven modeling approaches.

% Mention Koopman here (maybe) along with some of it's benefits.
Within the class of data-driven methods, deep learning or neural networks have been the primary choice for describing the input/output behavior of soft manipulators.
%This has enabled position control of several soft manipulators, but has not yet been shown effective under variable loading condition.
% yet been applied under variable loading conditions
% rarely / yet been utilized under variable loading conditions.
For instance, \citet{satheeshbabu2019open} used deep reinforcement learning to achieve open-loop position control of a soft manipulator comprised of fiber-reinforced actuators;
\citet{hyatt2019model} utilized a linearization of a neural network model and model predictive control to control the position of a bellows-actuated manipulator;
and \citet{thuruthel2018model} used a combination of a recurrent neural network and supervised reinforcement learning to achieve closed-loop control of a pneumatically-driven soft manipulator.
% \David{I suggest: ``This feedback controller was shown to work under various loading conditions.  
% Yet, to further improve system performance with regard to accuracy and speed it will become imperative to model and predict system behavior as a function of load.  
%To the best of the authors' knowledge, including such a load prediction and compensation has not been attempted in prior work.
% ''}
This controller was shown to compensate for disturbance such as end effector loading.
Moving forward, models that are able to predict system behavior as a function of load rather than treating them as a disturbance are required to improve system performance with regard to accuracy and speed.

% % Downside of nerual network models
% There are also fundamental downsides to using neural network models for control.
% Namely, building a neural network model requires solving a nonlinear optimization problem for which global convergence may not be guaranteed \cite{boyd2004convex}.
% Furthermore, utilizing a neural network model at run-time is non-trivial since the control input usually appear non-linearly within the computed model.

%% Introducing the Koopman operator and why it is great
An alternative to deep learning and artificial neural networks is a data-driven modeling approach based on Koopman operator theory. 
This approach yields a linear model that can be used for control.
These properties are in stark contrast to deep learning or neural network models, whose creation requires solving a nonlinear optimization problem for which global convergence may not be guaranteed \cite{boyd2004convex}, and in which the control input usually appears non-linearly.
% Koopman operator theory offers an alternative data-driven modeling approach that does not require solving a nonlinear optimization problem, and yields an explicit, control-oriented model.
The Koopman operator, on which this approach is based, describes the evolution of scalar-valued functions along trajectories of a nonlinear dynamical system, acting as a linear embedding of the nonlinear dynamics in an infinite-dimensional function space \cite{budivsic2012applied}\cite[Ch.~7]{brunton2019data}.
An approximation of this infinite-dimensional operator is identified via linear regression on input/output data \cite{bruder2019nonlinear, mauroy2016linear}, providing a linear model representation that is compatible with established linear control techniques \cite{Abraham-RSS-17, korda2018linear}.
A Koopman-based approach has been successfully used to control several robotic systems such as a Sphero SPRK,  \cite{Abraham-RSS-17}, a quadcopter \cite{abraham2019active}, and a robotic fish \cite{mamakoukas2019local}.
% Koopman has also been applied to soft contunuum 
Koopman-based system identification and control has also been successfully demonstrated on a soft continuum manipulator to control the 2D projection of its end-effector \cite{Vasudevan-RSS-19}.

% However, this previous work only controlled the position of the end effector in a 2D workspace, and did not take loading conditions into account.
% Therefore, the controller described in this work does not have the capacity to accommodate the loading conditions encountered in a typical manipulation task.

%% In this work, we present a data-driven modeling and control approach for soft robots that incorporates loads
\textbf{This paper presents a Koopman-based framework that explicitly incorporates loading conditions into the model to enable real-time control design.}
By incorporating loads into the model as states, our approach is able to estimate loading online via an observer within the control loop (see Fig.~1).
This observer infers the most likely value of the loading condition given a series of input/output measurements.
The knowledge that is gained in this process enables consistent control performance
% of a soft continuum manipulator
under a wide range of loading conditions.
In fact, the idea of estimating loading conditions has been explored for rigid-bodied robots in the past \cite{colome2013external,Funkhouser-RSS-19}.
By using our proposed Koopman-based approach, we are able to transfer these rigid-body results into the world of soft robots, which are subject to continuous deformation under load.

\textbf{Using this approach, we demonstrate real-time, fully autonomous control of a pneumatically actuated soft continuum manipulator}.
In several validation experiments our controller proves itself to be more accurate and more precise across various payloads than several other model-based controllers which do not incorporate loading.
%Our controller is also used to perform automated pick and place of objects of unknown mass.
To the best of the authors' knowledge, this paper presents the first implementation of a closed-loop controller that explicitly accounts for loading on a soft continuum manipulator and the first demonstration of autonomous pick and place of objects of unknown mass on a soft continuum manipulator.
% (see supplementary video attachment\footnote{\href{https://youtu.be/g2yRUoPK40c}{https://youtu.be/g2yRUoPK40c}}).

%% Outline
The rest of this paper is organized as follows:
Section \ref{sec:sysid} formally introduces the Koopman operator and describes how to construct models of nonlinear dynamical systems from data. 
Section \ref{sec:loadest} introduces a method for incorporating loading conditions into the model. 
Section \ref{sec:mpc} describes our Koopman-based model predictive controller and a method for estimating loading conditions online.
Section \ref{sec:experiments} describes the set of experiments used to evaluate the performance of our Koopman-based model predictive controller on a pneumatically actuated soft continuum manipulator, including trajectory following while carrying an unknown payload, and autonomously sorting objects by mass.
Section \ref{sec:discussion} discusses the results of these experiments and concludes the paper. 

\section{System Identification}    \label{sec:sysid}

%%%%%%%%%%%%%%%%%%%%%%%%%%%%%%%%%%%%%%%%%%%%%%%%%%%%%%%%%%%%%%%%%%
% System Identification
%%%%%%%%%%%%%%%%%%%%%%%%%%%%%%%%%%%%%%%%%%%%%%%%%%%%%%%%%%%%%%%%%%

% Overview of section
This section describes a system identification method to construct linear state space models of nonlinear controlled discrete dynamical systems from input/output data.
% Linearity is achieved by utilizing an approximation of the discrete-time Koopman operator.
In particular, rather than describing the evolution of a dynamical system's state directly, which may be a nonlinear mapping, the (linear) Koopman operator describes the evolution of scalar-valued functions of the state, which is a linear mapping in the infinite-dimensional space of all scalar-valued functions. 

%% Overview of the Koopman operator and how it represents dynamical systems
\subsection{The Kooman Operator}
\label{sec:koop_rep}

%% System representation in state space
Consider an input/output system governed by the following differential equation for the output:
\begin{align}
    \dot{y}(t) &= F ( y(t) , u(t) )
    % y(t) &= g( x(t) )
    \label{eq:nlsys}
\end{align}
where $y(t) \in Y \subset \Real^n$ is the output and $u(t) \in U \subset \Real^m$ is the input of the system at time $t \geq 0$, ${F}$ is a continuously differentiable function, and $Y$ and $U$ are compact subsets.
Denote by $\phi_\tau (y_0, u_0)$ the \emph{flow map}, which is the solution to \eqref{eq:nlsys} at time $\tau$ when beginning with the initial condition $y_0$ at time $0$ and a constant input $u_0$ applied for all time between $0$ and $\tau$.

%% System representation in the space of observables
The system can be lifted to an infinite-dimensional function space $\mathcal{F}$ composed of all square-integrable real-valued functions with compact domain $Y \times U \subset \Real^{n \times m}$.
Elements of $\mathcal{F}$ are called \emph{observables}.
In $\mathcal{F}$, the flow of the system is characterized by the set of Koopman operators $\koop_\tau : \mathcal{F} \to \mathcal{F}$, for each $\tau \geq 0$,
which describe the evolution of the observables ${f \in \mathcal{F}}$ along the trajectories of the system according to the following definition:
\begin{align}
    \koop_\tau f = f \circ \phi_\tau,      
    % && \forall f \in \F, t \geq 0
    \label{eq:koopman}
\end{align}
where $\circ$ indicates function composition.
A consequence of this definition is that for a specific time step $\tau$, the Koopman operator $\koop_\tau$ defines an infinite-dimensional linear discrete dynamical system that advances the value of an observable by $\tau$, 
% according the the dynamics of \eqref{eq:nlsys},
\begin{align}
    % f( y( k \tau + \tau ) , u ) &= \koop_\tau f( y( k \tau ) , u ) \hspace{20pt} , k \in \mathbb{N} \\
    f( y(t + \tau) , \tilde{u} ) &= 
    % f( \phi_\tau ( y(0) , u ) , u ) = 
    \koop_\tau f( y(t) , \tilde{u} )
    \label{eq:Uf}
\end{align}
where $\tilde{u}$ is a constant input over the interval $[t , t + \tau]$.
% This can be written using discrete notation as
% \begin{align}
%     f( y[k+1] , u[k] ) &= \koop_\tau f( y[k] , u[k] ).
% \end{align}
Since this is true for \emph{any} observable function $f$, the Koopman operator can be used to advance the output itself by applying it to the set of functions $\{ f_i : f_i (y(t) , \tilde{u}) = y_i (t) \}_{i=1}^n$, advancing their values according to \eqref{eq:Uf}, and stacking the results as a vector:
\begin{align}
    y( t + \tau ) &= \begin{bmatrix} \mathcal{K}_{\tau} f_1 \left( y (t) , \tilde{u} \right) & \cdots & \mathcal{K}_{\tau} f_n \left( y (t) , \tilde{u} \right) \end{bmatrix}^\top .
    \label{eq:ystep}
\end{align}
In this way, the Koopman operator provides an infinite-dimensional linear representation of a nonlinear dynamical system \cite{budivsic2012applied}.

%% Koopman sysid, i.e. EDMD
\subsection{Koopman-based System Identification}
\label{sec:koopid}

Since the Koopman operator is an infinite-dimensional object, we have to settle for its projection onto a finite-dimensional subspace, which can be represented as a matrix.
Using the Extended Dynamic Mode Decomposition (EDMD) algorithm \cite{williams2015data, mauroy2016linear, mauroy2017koopman}, we identify a finite-dimensional matrix approximation of the Koopman operator via linear regression applied to observed data.
The remainder of this subsection describes the mathematical underpinnings of this process.

Define ${\bar{\mathcal{F}} \subset \mathcal{F}}$ to be the subspace of $\mathcal{F}$ spanned by ${N>n+m}$ linearly independent basis functions
${ \{ \psi_i : \Real^{n \times m} \to \Real \}_{i=1}^N}$,
and define the \emph{lifting function} ${\psi : \Real^{n \times m} \to \Real^N}$ as:
\begin{align}
    % \psi(x) &:= \begin{bmatrix} \psi_1 (x) & \cdots & \psi_N (x) \end{bmatrix}^\top.
    \psi(y,u) &:= \begin{bmatrix} \psi_1(y,u) & \cdots & \psi_N (y,u) \end{bmatrix}^\top.
    \label{eq:lift}
\end{align}
Any observable $\bar{f} \in \bar{\mathcal{F}}$ can be expressed as a linear combination of the basis functions
\begin{align}
    \bar{f} &= \theta_1 \psi_1 + \cdots + \theta_N \psi_N
    \label{eq:fexpanded}
\end{align}
where each $\theta_i \in \Real$ is a constant.
Thus $\bar{f}$ evaluated at $y,u$ can be concisely expressed as
\begin{align}
    \bar{f}(y,u) &= \theta^\top \psi (y,u)
    \label{eq:fvec}
\end{align}
where ${\theta := [ \theta_1 \,  \cdots \, \theta_N ]^\top}$ acts as the \emph{vector representation} of $\bar{f}$.

Given this vector representation for observables, a linear operator on $\bar{\mathcal{F}}$ can be represented as an ${N \times N}$ matrix. 
We denote by $\bar{\koop}_\tau \in \Real^{N \times N}$ the approximation of the Koopman operator on $\bar{\mathcal{F}}$, which operates on observables via matrix multiplication:
\begin{align}
    \bar{\koop}_\tau \theta = \theta'
\end{align}
where $\theta , \theta'$ are each vector representations of observables in $\bar{\mathcal{F}}$.
%% Pulled straight from ICRA paper below this line
Our goal is to find a $\bar{\koop}_\tau$ that describes the action of the infinite dimensional Koopman operator $\koop_\tau$ as accurately as possible in the $L^2$-norm sense on the finite dimensional subspace $\bar{\mathcal{F}}$ of all observables.

For $\bar{\koop}_{\tau}$ to perfectly mimic the action of $\koop_\tau$ on any observable ${\bar{f} \in \bar{\mathcal{F}} \subset \mathcal{F}}$, according to \eqref{eq:koopman} the following should be true for all  $y \in Y$ and $u \in U$,
\begin{align}
    \bar{\koop}_\tau \bar{f}(y,u) &= \bar{f} \circ \phi_\tau(y,u) \label{eq:first} \\
    ( \bar{\koop}_\tau {\theta} )^\top {\psi}(y,u) &=
    {\theta}^\top {\psi} \circ \phi_\tau(y,u) \label{eq:second}\\
    \bar{\koop}_\tau^\top \psi(y,u) &= {\psi} \circ \phi_\tau(y,u),
    \label{eq:UbarEq}
\end{align}
where \eqref{eq:second} follows by substituting \eqref{eq:fvec} and \eqref{eq:UbarEq} follows since the result holds for all $\bar{f}$.
Since this is a linear equation, it follows that for a given ${y \in Y}$  and $u \in U$, solving \eqref{eq:UbarEq} for $\bar{\koop}_\tau$ yields the best approximation of $\koop_\tau$ on $\bar{\mathcal{F}}$ in the $L^2$-norm sense \cite{penrose1956best}:
\begin{align}
    \bar{\koop}_\tau = \left( {\psi}(y,u)^\top \right)^\dagger ( {\psi} \circ \phi_\tau(y,u) )^\top
    \label{eq:Uapprox}
\end{align}
where superscript $\dagger$ denotes the Moore-Penrose pseudoinverse.

%% How this is done on our system
To approximate the Koopman operator from a set of experimental data, we take $K$ discrete measurements in the form of so-called ``snapshots'' ${ (a[k] , b[k] , u[k]) }$ for each ${ k \in \{1,\ldots,K\} }$ where
% with sampling period $T_s$. We separate the data into a set of $K$ so-called ``snapshot pairs'' of the form ${ \{ a_k , b_k \} \in \Real^{n \times 2} }$ where
\begin{align}
    % & a[k] = \begin{bmatrix} y[k]) \\ u[k] \end{bmatrix} 
    % && b[k] = \begin{bmatrix} \phi_{T_s} (y[k] , u[k]) + \sigma[k] \\ u[k] \end{bmatrix}
    a[k] &:=  y[k] \\
    b[k] &:= \phi_{T_s} (y[k] , u[k]),
    \label{eq:ab}
\end{align}
$y[k]$ denotes the output corresponding to the $k^\text{th}$ measurement, $u[k]$ is the constant input applied between $a[k]$ and $b[k]$, and $T_s$ is the sampling period, which is assumed to be identical for all snapshots.
Note that consecutive snapshots do not have to be generated by consecutive measurements. 
% For our basis of $\bar{\mathcal{F}}$, we choose the basis of monomials of $x$ with total degree less than or equal to $w$, which implies ${N=(n+m+w)!/\left((n+m)!w!\right)}$ \cite[Section III]{mauroy2016linear}. 
We then lift all of the snapshots according to \eqref{eq:lift} and compile them into the following ${K \times N}$ matrices:
\begin{align}
    & \Psi_a := \begin{bmatrix} {\psi}(a[1] , u[1])^\top \\ \vdots \\  {\psi}(a[K] , u[K])^\top \end{bmatrix},
    && \Psi_b := \begin{bmatrix} {\psi}(b[1] , u[1])^\top \\ \vdots \\  {\psi}(b[K] , u[K])^\top \end{bmatrix}.
    \label{eq:Psi}
\end{align}
$\bar{\koop}_{T_s}$ is chosen so that it yields the least-squares best fit to all of the observed data, which, following from \eqref{eq:Uapprox}, is given by 
\begin{align}
    \bar{\koop}_{T_s} &:= \Psi_a^\dagger \Psi_b. \label{eq:Ubar}
\end{align}

%% Incorporating Delays
Sometimes a more accurate model can be obtained by incorporating delays into the set of snapshots. 
To incorporate these delays, we modify the snapshots to have the following form
\begin{align}
    & a[k] := \begin{bmatrix} y[k] \\ \vdots \\ y[k-d] \\ u[k-1] \\ \vdots \\ u[k-d] \end{bmatrix},
    % a[k] &= \left[ y[k]^\top , \ldots , y[k-d]^\top , u[k-1]^\top , \ldots , u[k-d]^\top \right]^\top \\
    && b[k] := \begin{bmatrix} \phi_{T_s} (y[k],u[k]) \\ \vdots \\ y[k-d+1] \\ u[k-1] \\ \vdots \\ u[k-d] \end{bmatrix}
    % b[k] &= \left[ \left( \phi_{T_s} (y[k],u[k]) + \sigma_k \right)^\top , \ldots , y[k-d+1]^\top , u[k-1] , \ldots , u[k-d] \right]^\top
    \label{eq:snapd}
\end{align}
where $d$ is the number of delays.
We then modify the domain of the lifting function such that ${ \psi : \Real^{\left( n+(n+m)d \right) \times m} \to \Real^{N} }$ to accommodate the larger dimension of the snapshots.
Once these snapshots have been assembled, the model identification procedure is identical to the case without delays.

% Optional: SVD on lifed data to find best N-dimensional state (this way we can use SUPER HUGE initial basis, and make it smaller)

\subsection{Linear Model Realization Based on the Koopman Operator}
\label{sec:linid}

For dynamical systems with inputs, we are interested in using the Koopman operator to construct discrete linear models of the following form
\begin{equation}
\begin{aligned}
    z[j+1] &= A z[j] + B u[j] \\
    y[j] &= C z[j]
    \label{eq:linSys}
\end{aligned}
\end{equation}
for each $j \in \mathbb{N}$, where $y[0]$ is the initial output, $z[0]$ is the initial state, and $u[j] \in \Real^m$ is the input at the $j^{\text{th}}$ step.
Specifically, we desire a representation in which the input appears \emph{linearly}, because models of this form are amenable to real-time, convex optimization techniques for feedback control design, as we describe in Section \ref{sec:mpc}.

% How to choose basis functions so that A,B matrices are inside of K
With a suitable choice of basis functions $\{ \psi_i \}_{i=1}^N$, $\bar{\koop}_{T_s}$ can be constructed such that it is decomposable into a linear system representation like \eqref{eq:linSys}.
% has the linear system matrices $A,B$ embedded within.
One way to achieve this is to define the first $N-m$ basis functions as functions of the output only, and the last $m$ basis functions as indicator functions on each component of the input,
\begin{align}
    \psi_i(y,u) &= g_i(y), \quad \forall i \in \{1,\ldots, N-m\}  \label{eq:gi} \\
    \psi_i(y,u) &= u_i, \quad \forall i \in \{i = N-m+1, \ldots, N\} \label{eq:ui}
\end{align}
where $g_i : \Real^n \to \Real$ and $u_i$ denotes the $i^{\text{th}}$ element of $u$.
This choice ensures that the input only appears in the last $m$ components of $\psi(y,u)$, and an $N-m$ dimensional state can be defined as $z = g( y ) \in \Real^{N-m}$, where ${ g : \Real^{n} \to \Real^{N-m} }$ is defined as
\begin{align}
    g( y ) &:= \begin{bmatrix} g_1 (y) & \cdots & g_{N-m} (y) \end{bmatrix}^\top.
\end{align}

Following from \eqref{eq:Ubar}, the transpose of $\bar{\koop}_{T_s}$ is the best transition matrix between the elements of the lifted snapshots in the $L^2$-norm sense.
This implies that given the lifting functions defined in \eqref{eq:gi} and \eqref{eq:ui}, $\bar{\koop}_{T_s}$ is the minimizer to
\begin{align}
    \underset{\check{\koop}}{\min} 
    \sum_{k=1}^K
    \left\lVert 
    {\check{\koop}}^\top \begin{bmatrix} g( a[k] ) \\ u[k] \end{bmatrix} - \begin{bmatrix} g( b[k] ) \\ u[k] \end{bmatrix}
    \right\rVert_2^2.
    \label{eq:min-K}
\end{align}
Also note that given ${ z = g( y ) }$ as the state of our linear system, the best realizations of $A$ and $B$ in the $L^2$-norm sense 
% based on our data 
are the minimizers to
\begin{align}
    \underset{\check{A}, \check{B}}{\min} 
    \sum_{k=1}^K
    \left\lVert 
    \check{A} g(a[k]) + \check{B} u[k] - g(b[k])
    \right\rVert_2^2 .
    \label{eq:min-AB}
\end{align}
Therefore, by comparing \eqref{eq:min-K} and \eqref{eq:min-AB}, one can confirm that $A$ and $B$ are embedded in $\bar{\koop}_{T_s}$ and can be isolated by partitioning it as follows:
\begin{align}
    \bar{\koop}_{T_s}^\top &= 
    \begin{bmatrix} 
        A_{(N-m) \times (N-m)} &
        B_{(N-m) \times m} \\
        O_{m \times (N-m)} &
        I_{m \times m}
    \end{bmatrix}
    \label{eq:AB}
\end{align}
where $I$ denotes an identity matrix, $O$ denotes a zero matrix, and the subscripts denote the dimensions of each matrix.

We can also define the first $n$ basis functions as indicator functions on each component of the output, i.e.
\begin{align}
    g_i(y) &= y_i \quad \forall i \in \{1, \ldots, n \}  \label{eq:psi-1-n}.
\end{align}
Then, $C$ is defined as the matrix which projects the first $n$ elements of the state onto the output-space,
\begin{align}
    C &= \begin{bmatrix} I_{n \times n} & O_{n \times (N-n)} \end{bmatrix}.
    \label{eq:C}
\end{align}

%%%%%%%%%%%%%%%%%%%%%%%%%%%%%%%%%%%%%%%%%%%%%%%%%%%%%%%%%%%%%%%%%%
% Online Load Estimation
%%%%%%%%%%%%%%%%%%%%%%%%%%%%%%%%%%%%%%%%%%%%%%%%%%%%%%%%%%%%%%%%%%
\subsection{Incorporating Loading Conditions Into the Model}
\label{sec:loadest}

% Why do we need to consider loading differently than other states?
For robots that interact with objects or their environment, understanding the effect of external loading on their dynamics is critical for control.
These loading conditions alter the dynamics of a system, but are generally not directly observable.
This poses a challenge for model-based control, which relies on an accurate dynamical model to choose suitable control inputs for a given task.

% what do we want? Load estimation and control
We desire a way to incorporate loading conditions into our dynamic system model and to estimate them online.
We can achieve this by including them within the states 
% \David{I was a bit puzzled by the wording ``states'' here.  My thinking was that your states $z$ are lifted states and that $y$ are outputs. So in a sense, are $w$ outputs???  Not sure, maybe I'd be less confused when you say ``including a parameterization of the load within the lifted states'', since with the current wording I started to think that there are `non-lifted' states, which is not true } 
of our Koopman-based lifted system model, and then constructing an online observer to estimate their values.
This strategy utilizes the underlying model to infer the most likely value of the loading conditions given past input/output measurements.

Let $w \in \Real^{p}$ be a parametrization of loading conditions.
For example, $w$ might specify the mass at the end effector of a manipulator arm.
We incorporate $w$ into the state $z$ using a new lifting function ${ \gamma : \Real^{n \times p} \to \Real^{(N-m)(p+1)} }$,
which accepts $w$ as a second input and is defined as:
\begin{align}
    z &= \gamma(y,w) = 
        \begin{bmatrix}
            g(y) \\ g(y) w_1 \\ \vdots \\ g(y) w_p
        \end{bmatrix}
        = 
        \underbrace{
        \begin{bmatrix}
            g(y) & \cdots & 0 \\ \vdots & \ddots & \vdots \\
            0 & \cdots & g(y)
        \end{bmatrix}
        }_{\Gamma(y)}
        % \underbrace{
        \begin{bmatrix}
            1 \\ w
        \end{bmatrix}
        % }_{W}
        \label{eq:gamma}
\end{align}
where ${ \Gamma(y) \in \Real^{\left((N-m)(p+1)\right) \times (p+1)} }$ is the matrix formed by diagonally concatenating $g(y)$, ${p+1}$ times, 
% $W = [1 , w^\top]^\top$, 
and $w_i$ denotes the $i^\text{th}$ element of $w$.
Note that because this lifting function requires the loading condition $w$ as an input, it must also be included in the snapshots ${ (a[k] , b[k] , u[k] , w[k] )}$ to construct a Koopman model that accounts for loading. 

% How to estimate w
Although $w$ is not measured directly, its value can be inferred based on the system model and past input-output measurements.
We construct an observer that estimates the value of $w$ at the $j^\text{th}$ timestep by solving a linear least-squares problem using data from the $N_w$ previous timesteps.
Notice that the output at the $j^\text{th}$ timestep $y[j]$ can be expressed in terms of the input $u[j-1]$, the output $y[j-1]$, and the load $w[j-1]$ at the previous timestep by combining the system model equations of \eqref{eq:linSys} and then substituting \eqref{eq:gamma} for $z[j-1]$, 
\begin{align}
    y[j] &= C A z[j-1] + C B u[j-1] \\
    &= C A \Gamma( y[j-1] ) \begin{bmatrix} 1 \\ w[j-1] \end{bmatrix} + C B u[j-1].
    \label{eq:y-from-zu}
\end{align}
Solving for the best estimate of $w[j-1]$ in the $L^2$-norm sense, denoted $\hat{w}[j-1]$ yields the following expression,
\begin{align}
    \begin{bmatrix} 1 \\ \hat{w}[j-1] \end{bmatrix}
    &= \left( C A \Gamma( y[j-1] ) \right)^\dagger \left( y[j] - C B u[j-1] \right)
    \label{eq:what-jm1}
\end{align}
where $^\dagger$ denotes the Moore-Penrose psuedoinverse.

Under the assumption that the loading is equal to some constant $\tilde{w}$ over the previous $N_w$ time steps, i.e. $w[i] = \tilde{w}$ for $i=j-N_w,\ldots,j$, we can similarly find the best estimate over all $N_w$ timesteps.
Since this estimate is based on more data it should be more accurate and more robust to noisy output measurements.
We define the following two matrices,
\begin{align}
    & \Lambda_A =
    \begin{bmatrix} 
        C A \Gamma( y[j-1] ) \\ \vdots \\ C A \Gamma( y[j-N_w] )
    \end{bmatrix} \\
    & \Lambda_B =
    \begin{bmatrix} 
        y[j] - C B u[j-1] \\ \vdots \\ y[j-N_w +1] - C B u[k-N_w]
    \end{bmatrix}
    \label{eq:Lambda-matrices}
\end{align}
% and \eqref{eq:y-from-zu} dictates that the following equation should hold
% \begin{align}
%     \Lambda_A W &= \Lambda_B 
% \end{align}
Then, following from \eqref{eq:what-jm1}, the best estimate for $\bar{w}$ over the past $N_w$ timesteps in the $L^2$-norm sense, denoted $\hat{w}$, is given by
\begin{align}
    \begin{bmatrix} 1 \\ \hat{w} \end{bmatrix} &= \Lambda_A^\dagger \Lambda_B,
    \label{eq:what}
\end{align}
% which provides the loading $w$ since it is embedded within $W$.
where $^\dagger$ again denotes the Moore-Penrose psuedoinverse.

%%%%%%%%%%%%%%%%%%%%%%%%%%%%%%%%%%%%%%%%%%%%%%%%%%%%%%%%%%%%%%%%%%
% Control 
%%%%%%%%%%%%%%%%%%%%%%%%%%%%%%%%%%%%%%%%%%%%%%%%%%%%%%%%%%%%%%%%%%
\section{Control}
\label{sec:mpc}

A system model enables the design of model-based controllers that leverage model predictions to choose suitable control inputs for a given task.
In particular, model-based controllers can anticipate future events, allowing them to optimally choose control inputs over a finite time horizon.
A popular model-based control design technique is model predictive control (MPC), wherein one optimizes the control input over a finite time horizon, applies that input for a single timestep, and then optimizes again, repeatedly \cite{rawlings2009model}.
For linear systems, MPC consists of iteratively solving a convex quadratic program (QP).
Importantly, this is also the case for Koopman-based linear MPC control, 
wherein one solves for the optimal sequence of control inputs over a receding prediction horizon \cite[Eq.~23]{korda2018linear}.

The predictions of this Koopman-based controller depend on the estimate of the loading conditions $\hat{w}$.
This estimate must be periodically updated using the method described in Section~\ref{sec:loadest}, but for systems with relatively stable loading conditions, it is computationally inefficient to compute a new estimate at every time step.
Therefore, we define a load estimation update period $N_e$ as the number of time steps to wait between load estimations.
Increasing $N_e$ will likely increase the accuracy of each load estimate, but will also reduce responsiveness to changes in the loading conditions. To balance accuracy with responsiveness, we update $\hat{w}$ every $N_e$ time steps by setting it equal to the average of the new load estimate and the previous $N_r$ load estimates, where $N_r$ is another user defined constant.
Algorithm \ref{alg:mpc} summarizes the closed-loop operation of this Koopman-based MPC controller with these periodic load estimation updates.

%% MPC algorithm
\begin{algorithm}[t]
\SetAlgoLined
\KwIn{ Prediction horizon: $N_h$ \\
    %   \hspace{32pt} Cost matrices: $G_i , H_i , g_i , h_i$ for  $i = 0 , ... ,N_h$ \\
    %   \hspace{32pt} Constraint matrices: $E_i , F_i , b_i$ for $i = 0 , ... ,N_h$ \\
       \hspace{32pt} Model matrices: $A , B , C$}
\For{ $k = 0 , 1 , 2 , ... $}{
\eIf{$k \mod N_e = 0$}{
    Estimate $\hat{w}'[j]$ via \eqref{eq:what} \\
    $\hat{w}[k]$ = mean($ \hat{w}'[j] , \hat{w}[k-1] , \ldots , \hat{w}[k-N_{r}] $) \\
    $j = j+1$
   }{
   $\hat{w}[k] = \hat{w}[k-1]$
  }
% \textbf{Step 1:} Set $z[0] = \gamma ( y[k] , \hat{w}[k] )$ \Dan{modify} \\
\textbf{Step 1:} Solve QP to find optimal input $(u[i]^*)_{i=0}^{N_h}$ \\
\textbf{Step 2:} Set $u[k] = u[0]^*$ \\
\textbf{Step 3:} Apply $u[k]$ to the system
}
 \caption{Koopman MPC with Load Estimation }
 \label{alg:mpc}
\end{algorithm}
\section{Experiments}  \label{sec:experiments}

%% Introduction
This section describes the soft continuum manipulator and the set of experiments used to demonstrate the efficacy of the modeling and control methods described in Sections \ref{sec:sysid} and \ref{sec:mpc}.
Footage from these experiments is included in a supplementary video file\footnote{\href{https://youtu.be/g2yRUoPK40c}{https://youtu.be/g2yRUoPK40c}}.
% and the code used to construct Koopman-based models and controllers from data can be found in a publicly accessible repository\footnote{https://github.com/ramvasudevan/soft-robot-koopman-loaded}.\Dan{Need to update repository} \Ram{maybe don't cite us...?}

%%%%%%%%%%%%%%%%%%%%%%%%%%%%%%%%%%%%%%%%%%%%%%%%%%%%%%%%%%%%%%%%%%%%%%%%%%%%
% Description of robot arm and system identification
%%%%%%%%%%%%%%%%%%%%%%%%%%%%%%%%%%%%%%%%%%%%%%%%%%%%%%%%%%%%%%%%%%%%%%%%%%%%
\subsection{System Identification of Soft Robot Arm}
\label{sec:sysid-arm}

% figure: robot schematic
\begin{figure}
    \centering
    \includegraphics[width=\linewidth]{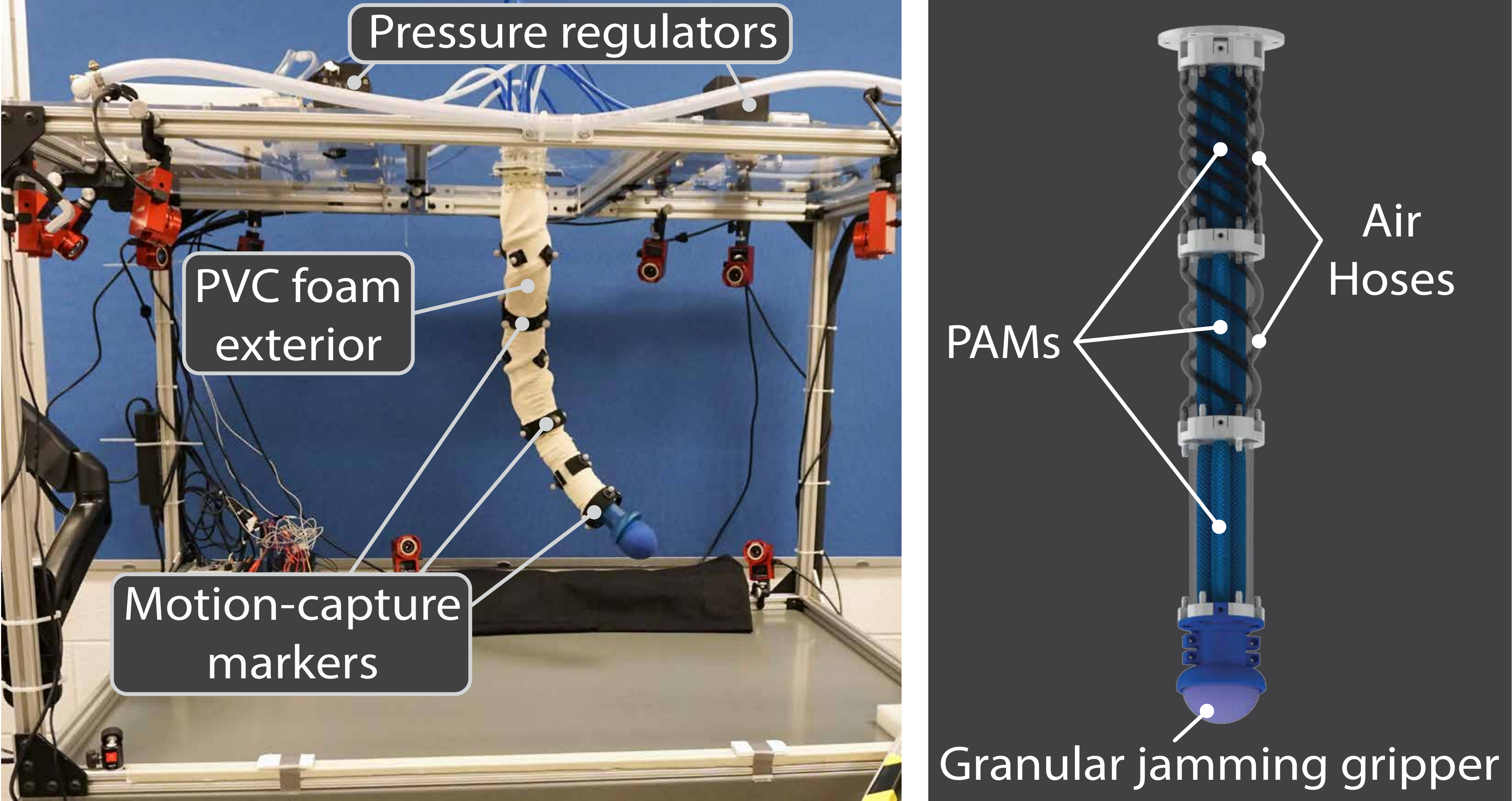}
    \caption{The soft robot arm consists of three bending sections, each actuated by three pneumatic artificial muscles (PAMs). The actuators are surrounded by a sleeve of flexible PVC foam, and pressurized air is supplied to the actuators via air hoses that wind around the exterior. The end effector consists of a granular jamming vacuum gripper \cite{amend2012positive}, which is connected to a vacuum pump by a hose that runs along the interior of the arm.} 
    \label{fig:robot-schematic}
\end{figure}

To validate the modeling and control approach described in the previous section, we applied it to a soft robot arm capable of picking-up objects and moving its end effector in three-dimensional space. 
The robot, shown in Fig.~\ref{fig:robot-schematic}, is 70~cm long and has a diameter of 6~cm.
It is made up of three pneumatically actuated bending sections and an end effector comprised of a granular jamming vacuum gripper \cite{amend2012positive}.
Each section is actuated by three pneumatic artificial muscles (PAMs) \cite{tondu2012modelling} which are adhered to a central spine consisting of an air hose encased in flexible PVC foam tubing.
Another much larger sleeve of flexible PVC foam surrounds the actuators, which serves to dampen high frequency oscillations and make the body of the arm softer overall.
The air pressure inside the actuators is regulated by $9$ Enfield TR-010-g10-s pneumatic pressure regulators that accept ${0-10}$V command signals corresponding to pressures of approximately ${0 - 275}$~kPa, and are connected to the actuators by air hoses that wrap around the outside of the foam sleeves.
The exterior of the arm is covered in retro-reflective markers which are tracked using a commercial OptiTrack motion capture system.

% Stochasticity
We quantified the stochastic behavior of our soft robot system by observing the variations in output from period-to-period under sinusoidal inputs  with a period of 10 seconds and a sampling time of $T_s = 0.083$ seconds with a zero-order-hold between samples.
Over 60 periods, the trajectory of the end effector deviated from the mean trajectory by an average of 9.45~mm and with a standard deviation of 7.3~mm.
This inherent stochasticity limits the tracking performance of the system, independent of the employed controller.

% input and output
For the purposes of constructing a dynamic model for the arm, the input was chosen to be the command voltages into the 9 pressure regulators and at each instance in time was restricted to $[0,10]^{9}$. 
The output was chosen to live in $\Real^9$ and corresponds to the positions of the ends of each of the 3 bending sections in Cartesian coordinates with the last 3 coordinates corresponding to the end effector position.
The parametrization of the loading condition lives in $\Real_{+}$ and is chosen to be the mass of the object held by the gripper.

% How data was collected
Data for constructing models was collected over $49$ trials lasting approximately $10$ minutes each.
A randomized ``ramp and hold'' type input and a load from the set $\{ 0,50,100,150,200,250,300 \}$ grams was applied during each trial to generate a representative sampling of the system's behavior over its entire operating range.
% To generate a random input, a matrix $\Upsilon \in [0,10]^{9\times 1000}$ of uniformly distributed random numbers between $0$ and $10$ was generated to be used as an input lookup table.
% Each control input was varied between elements in consecutive columns of the table over a transition period $T_u$, with a time offset of $T_u / 9$ between each of the nine control signals, 
% \begin{align}
%     u_i [k] &= \frac{(\Upsilon_{i,j+1} - \Upsilon_{i,j})}{T_u} \left( k T_s + \frac{(i-1) T_u}{3} \right) + \Upsilon_{i,j}
%     \label{eq:input}
% \end{align}
% where $j = \text{floor}\left( {k T_s} / {T_u} \right)$ is the current index into the lookup table at time $t$, ${ T_s = 0.083 }$ seconds is the sampling time, and there is a zero-order-hold between samples. 
% All trials were conducted using one of the transition periods $T_u$ in the set $\{0.5 , 1 , 2 , 3 \}$ seconds.

% Models were build from this data
Three models were fit from the data: a linear state-space model using the subspace method \cite{van2012subspace}, a linear Koopman model that \emph{does not} take loading into account using the approach described in Section~\ref{sec:koopid}, and a linear Koopman model that \emph{does} incorporate loading using the approach from Section~\ref{sec:loadest}.
Each of theses models was fit using the same set of $325,733$ randomly generated data points just once, independent of any specific task.

The linear state-space model provides a baseline for comparison and was identified using the MATLAB System Identification Toolbox \cite{MATLAB:2017}.
This model is a 9 dimensional linear state-space model expressed in observer canonical form.

The first Koopman model (without loading) was identified on a set of ${ K = 325,732 }$ snapshot pairs $\{ a[k] , b[k] , u[k] \}_{k=1}^K$ that incorporate a single delay ${d = 1}$:
\begin{align}
    a[k] &= \begin{bmatrix} y[k]^\top & y[k-1]^\top & u[k-1]^\top \end{bmatrix}^\top 
    \label{eq:real-a} \\
    b[k] &= \begin{bmatrix} \left( \phi_{T_s} (y[k]) + \sigma[k] \right)^\top & y[k]^\top & u[k]^\top \end{bmatrix}^\top.
    \label{eq:real-b}
\end{align}
Note that the dimension of each snapshot is ${ 2n+m = 2(9)+9 = 27 }$ due to the inclusion of the delay, 
and we denote by $y^d[k] \in \Real^{27}$ one of these outputs which has delays included at some time $k$.
The the lifting function ${ \psi : \Real^{27 \times 9} \to \Real^{111} }$ was defined as
\begin{align}
    \psi(y^d[k],u[k]) &= \begin{bmatrix} g(y^d[k]) \\ u[k] \end{bmatrix} 
    \label{eq:real-lift-1}
\end{align}
where the range of $g$ has dimension ${ N = 102 }$, ${ g_i(y^d[k]) = y^d_i[k] }$ for $i = 1,...,27$, and the remaining 75 basis functions $\{ g_i : \Real^{27} \to \Real \}_{i=28}^{102}$ are polynomials of maximum degree 2 that were selected 
% A collection of ${ N = 102 }$ polynomials of maximum degree 2 were selected as basis functions 
%
by evaluating the snapshot pairs on the set of all monomials of degree less then or equal to 2, then performing principle component analysis (PCA) \cite[Ch.~1.5]{brunton2019data} to identify a reduced set of polynomials that can still explain at least  99\% of this lifted data.

The second Koopman model (with loading) was identified on the same set of snapshot pairs as the first model, but with the loading included $\{ a[k] , b[k] , u[k] , w[k] \}_{k=1}^K$.
The lifting function ${ \psi : \Real^{27 \times 9} \to \Real^{231} }$ was defined as,
\begin{align}
    \psi(y^d[k],u[k],w[k]) &= \begin{bmatrix} \gamma(y^d[k],w[k]) \\ u[k] \end{bmatrix} 
    = \begin{bmatrix} g(y^d[k]) \\ g(y^d[k]) w[k] \\ u[k] \end{bmatrix}
    \label{eq:real-lift-2}
\end{align}
where the range of $g$ has dimension ${ N = 111 }$, $g_i(y^d[k]) = y^d_i[k]$ for $i = 1,...,27$, and the remaining 84 basis functions $\{ g_i : \Real^n \to \Real \}_{i=28}^{111}$ are polynomials of maximum degree 2 that were selected using the same PCA method described in the previous paragraph..
% A collection of ${ N = 111 }$ polynomials of maximum degree 2 were selected as basis functions using the same PCA method described in the previous paragraph.
% $a[k]$ and $b[k]$ were defined the same as in \label{eq:real-a} \label{eq:real-b} incorporated a single delay ${d=1}$ defined by \label{eq:real-1}

%%%%%%%%%%%%%%%%%%%%%%%%%%%%%%%%%%%%%%%%%%%%%%%%%%%%%%%%%%%%%%%%%%%%%%%%%%%%
% Description of Controllers
%%%%%%%%%%%%%%%%%%%%%%%%%%%%%%%%%%%%%%%%%%%%%%%%%%%%%%%%%%%%%%%%%%%%%%%%%%%%
\subsection{Description of Controllers}
\label{sec:mpc-arm}

% Description of task
Three model predictive controllers were constructed using the data-driven models described in the previous section.
Each controller uses one of the identified models to compute online predictions and is denoted by an abbreviation specifying which model,
\begin{itemize}
    \item L-MPC: Uses the linear state-space model
    \item K-MPC: Uses the Koopman model without loading
    \item KL-MPC: Uses the Koopman model with loading
\end{itemize}
% All three controllers computed inputs over a $12$-step prediction horizon ($N_h = 12$) online at a rate of $12$~Hz.
All three controllers solve a quadratic program at each time step using the Gurobi Optimization software \cite{gurobi}.
They run in closed-loop at $12$~Hz, feature an MPC horizon of $1$ second ($N_h = 12$), and a cost function that penalizes deviations of the position of the end effector from a reference trajectory over the prediction horizon.

\subsection{Experiment 1: Trajectory Following with Known Payload}
\label{sec:exp1}

\begin{figure*}[t]
    \centering
    \parbox[][][c]{0.32\linewidth}{\centering L-MPC}
    \parbox[][][c]{0.32\linewidth}{\centering K-MPC}
    \parbox[][][c]{0.32\linewidth}{\centering KL-MPC}
    \includegraphics[width=0.32\linewidth]{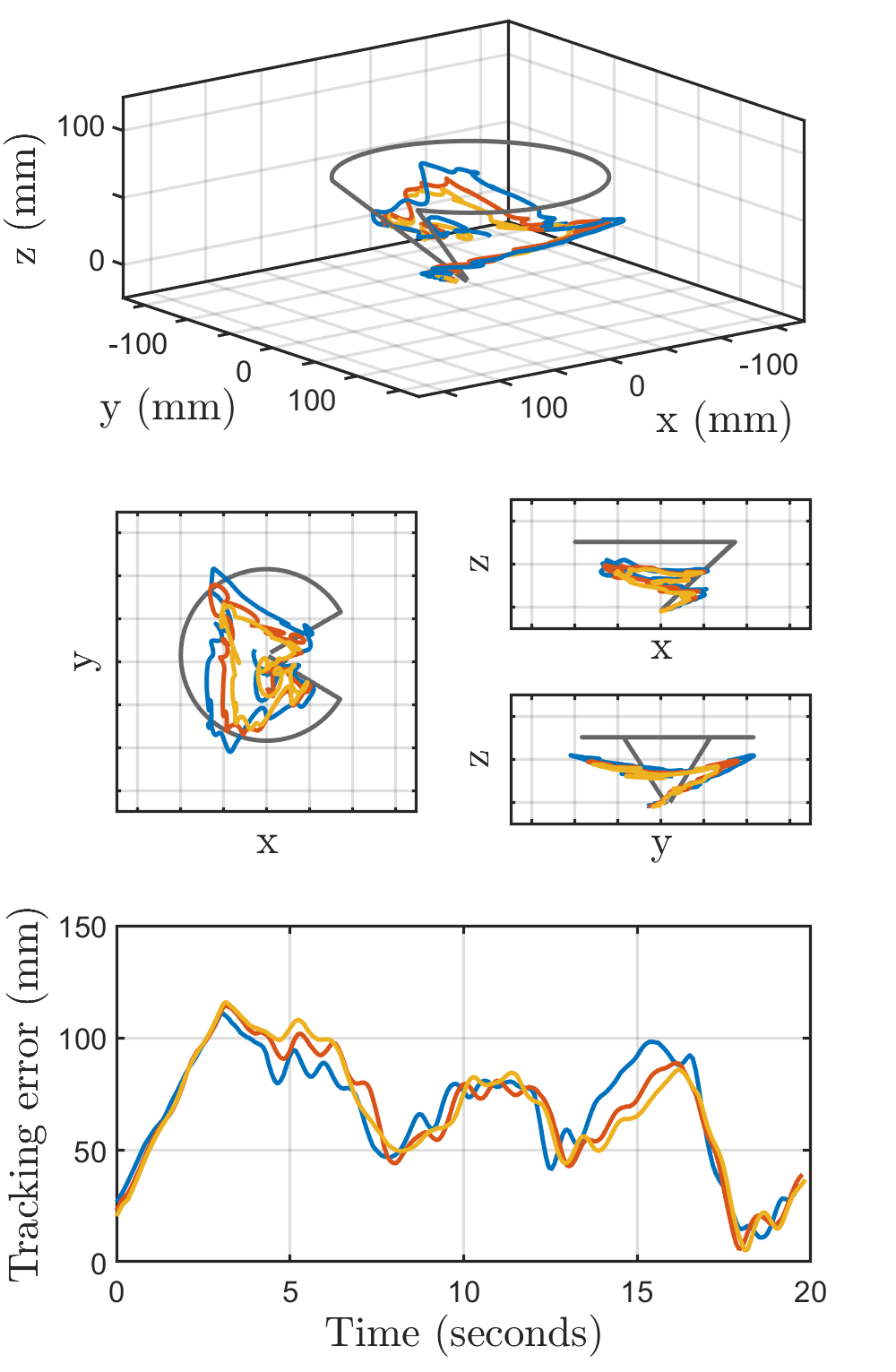}
    \includegraphics[width=0.32\linewidth]{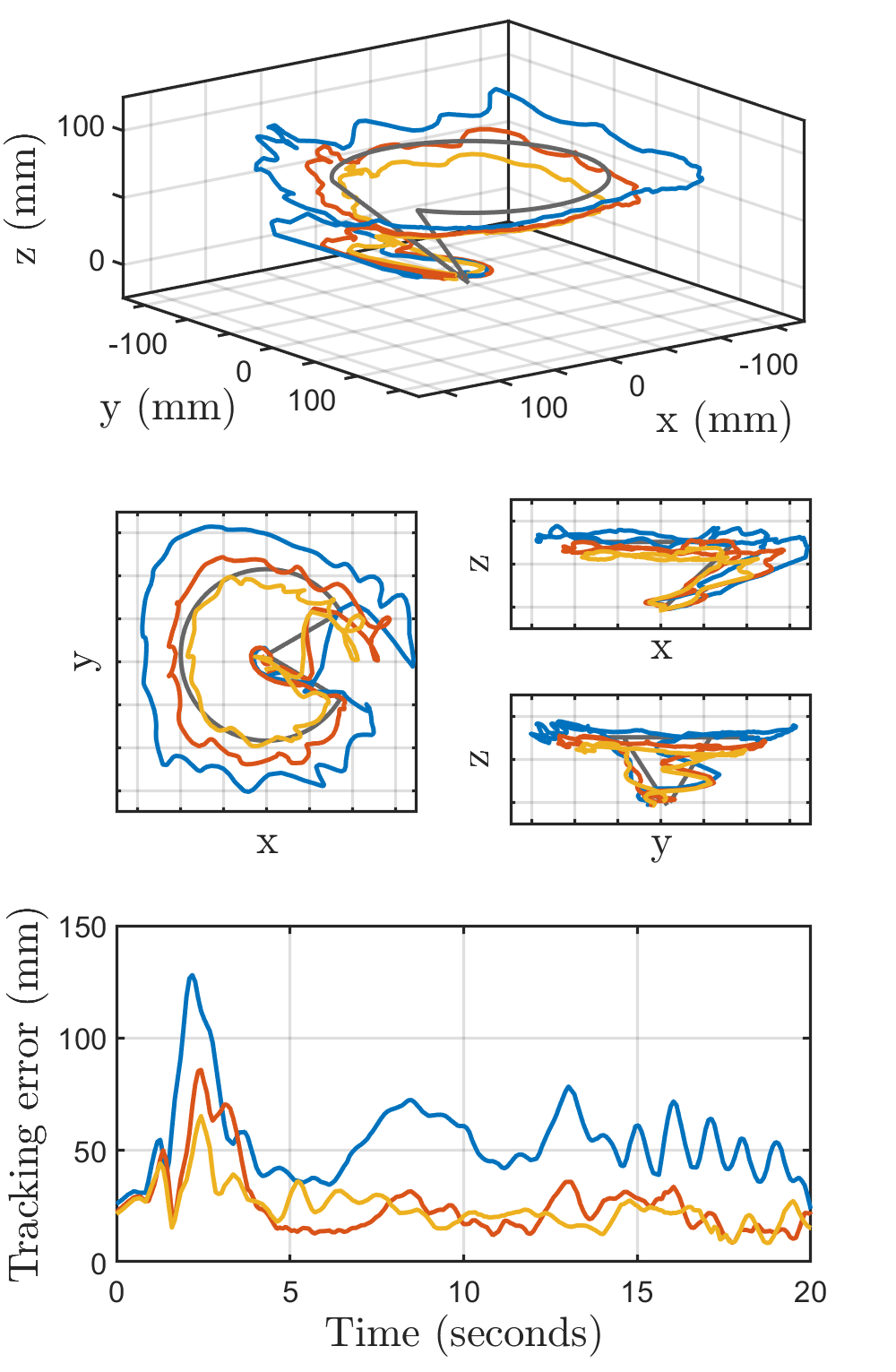}
    \includegraphics[width=0.32\linewidth]{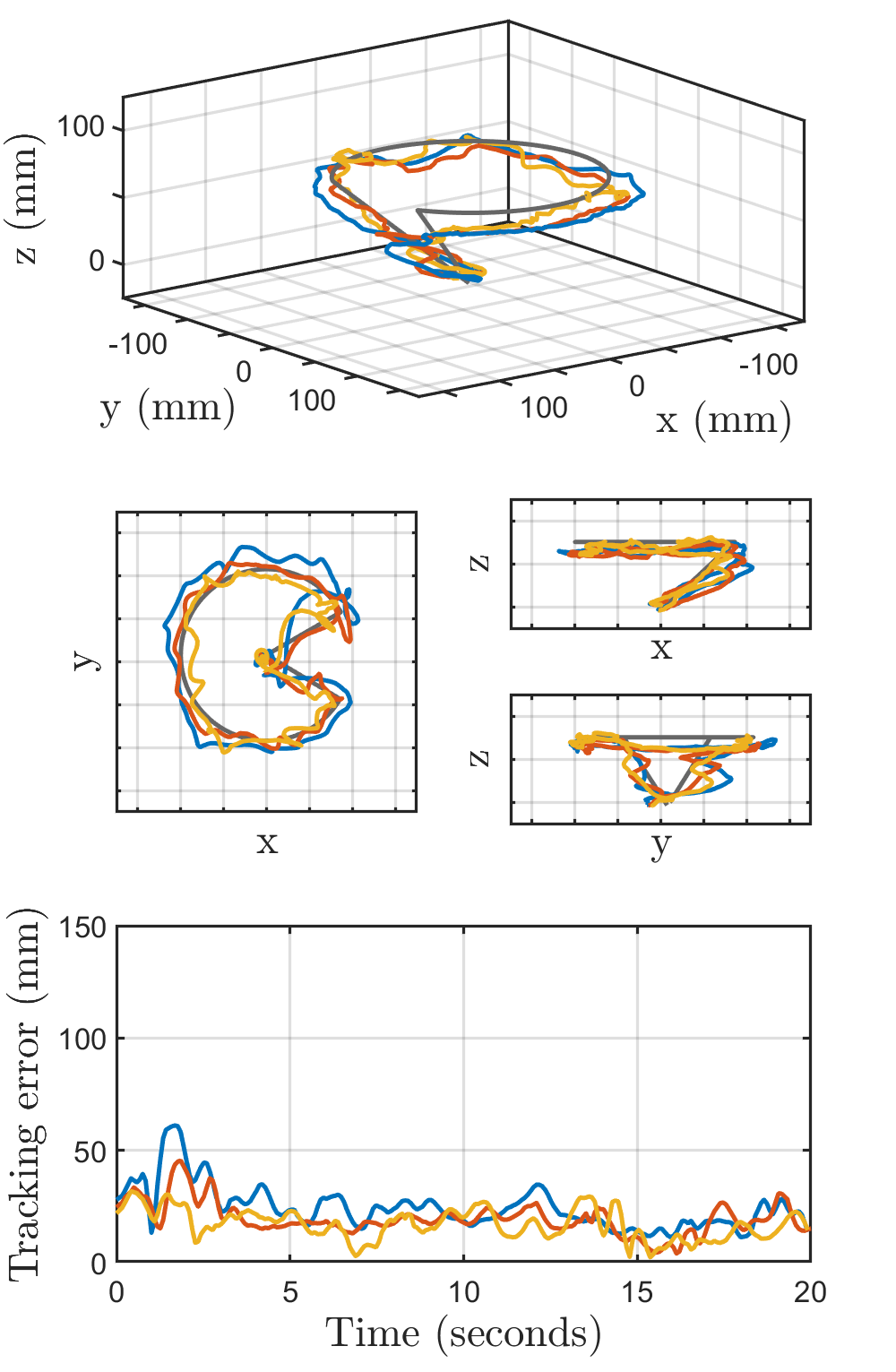}
    \caption{Experiment 1 Results: The end effector trajectories for the L-MPC (left), K-MPC (center), and KL-MPC (right) controllers when the true value of the payload is known. Trajectories corresponding to a payload of 25g are shown in blue, trajectories with a payload of 125g are shown in red, trajectories with a payload of 225g are shown in yellow, and the reference trajectory is shown in grey.}
    \label{fig:control-known-load}
\end{figure*}

% To evaluate controller performance independently/irrespective of the accuracy of online load estimations,
We first evaluated the relative performance of the three controllers when the payload at the end effector is known.
% That is, $\hat{w}$ in \eqref{eq:mpc} is set equal to the actual payload.
With this information given, the manipulator is tasked with moving the end effector along a three-dimensional reference trajectory lasting 20~seconds.
Six trials were completed for payloads of 25, 75, 125, 175, 225, and 275 grams.
The actual paths traced out by the end effector and the tracking error over time for 3 of the trials are displayed in Fig.~\ref{fig:control-known-load}, and the RMSE tracking error for all 6 trials is compiled in Table~\ref{tab:comparison}.
It should be noted that only the KL-MPC controller is capable of actually utilizing knowledge of the payload, since the other 2 controllers are based on models that do not incorporate loading conditions.

%% TABLE: RMSE results table
\begin{table}
    \rowcolors{2}{white}{gray!25}
    \setlength\tabcolsep{5pt} % default value: 6pt
    \centering
    \caption{Experiment 1: RMSE (mm) over entire trial}
    \begin{tabular}{|c|c|c|c|c|c|c|c|c|}
        \hline
        \rowcolor{white} 
        & \multicolumn{6}{c |}{\textbf{Payloads (grams)}} & & \textbf{Std.} \\
        \hhline{~------~~} \rowcolor{white}
        \multirow{-2}{*}{\textbf{Controller}} & 25 & 75 & 125 & 175 & 225 & 275 & \multirow{-2}{*}{\textbf{Avg.}} & \textbf{Dev.} \\
        \hline
        % RESULTS FOR ROBOT A
        L-MPC   &  73.0 & 72.9 & 72.6 & 71.9 & 72.3 & 74.3 & 72.8 & 0.8 \\
        K-MPC   &  55.4 & 33.9 & 29.5 & 20.0 & 24.8 & 27.8 & 31.9 & 12.4 \\
        KL-MPC  &  \textbf{26.1} & \textbf{23.7} & \textbf{20.6} & \textbf{19.5} & \textbf{18.2} & \textbf{20.4} & \textbf{21.4} & \textbf{2.9} \\
        \hline
    \end{tabular}
    \label{tab:comparison}
\end{table}

%%%%%%%%%%%%%%%%%%%%%%%%%%%%%%%%%%%%%%%%%%%%%%%%%%%%%%%%%%%%%%%%%%%%%%%%%%%%
% Experiment 2: Online load estimation
%%%%%%%%%%%%%%%%%%%%%%%%%%%%%%%%%%%%%%%%%%%%%%%%%%%%%%%%%%%%%%%%%%%%%%%%%%%%
\subsection{Experiment 2: Online Estimation of Unknown Payload}
\label{sec:exp2}

\begin{figure}
    \centering
    \includegraphics[width=\linewidth]{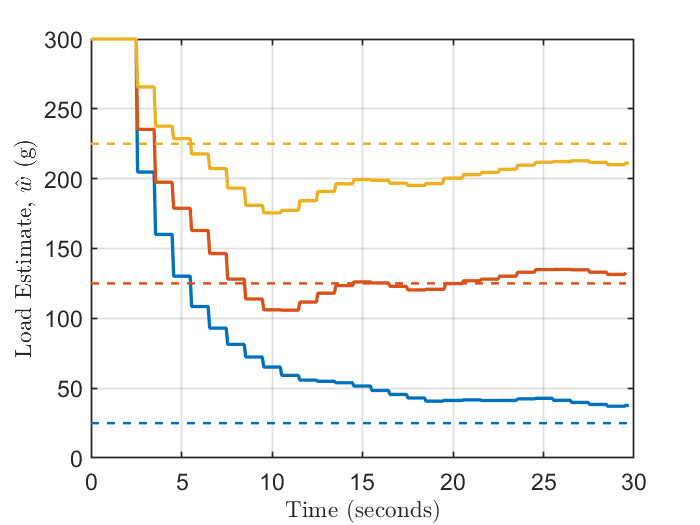}
    \caption{Experiment 2 Results: Online payload estimation under random inputs using the method described in Section~\ref{sec:loadest}. Three trials are shown for payloads of 25g, 125g, and 225g, with the actual payload used for each trial marked by a dotted line, and the payload estimate marked a solid line.
    Results for the 25g payload are shown in blue, results for the 125g payload are shown in red, and results for the 225g payload are shown in yellow.}
    \label{fig:load-estimation}
\end{figure}

We evaluated the performance of the online load estimation method described in Sections \ref{sec:loadest} and \ref{sec:mpc}
under randomized ``ramp and hold'' type inputs and a sampling time of $T_s = 0.083$ seconds.
New estimates were calculated  every ${N_e = 12}$ timesteps by solving \eqref{eq:what} using measurements from the previous ${ N_w = 30 }$ timesteps, and $\hat{w}$ was computed by averaging over the most recent ${ N_r = 360 }$ estimates.
Three trials were conducted with payloads of 25, 125, and 225 grams, none of which were in the set of payloads used for system identification, and the results are displayed in Fig.~\ref{fig:load-estimation}

%%%%%%%%%%%%%%%%%%%%%%%%%%%%%%%%%%%%%%%%%%%%%%%%%%%%%%%%%%%%%%%%%%%%%%%%%%%%
% Experiment 3: Trajectory following with unknown load
%%%%%%%%%%%%%%%%%%%%%%%%%%%%%%%%%%%%%%%%%%%%%%%%%%%%%%%%%%%%%%%%%%%%%%%%%%%%
\subsection{Experiment 3: Trajectory Following with Unknown Payload}
\label{sec:exp3}

\begin{figure}
    \centering
    \includegraphics[width=\linewidth]{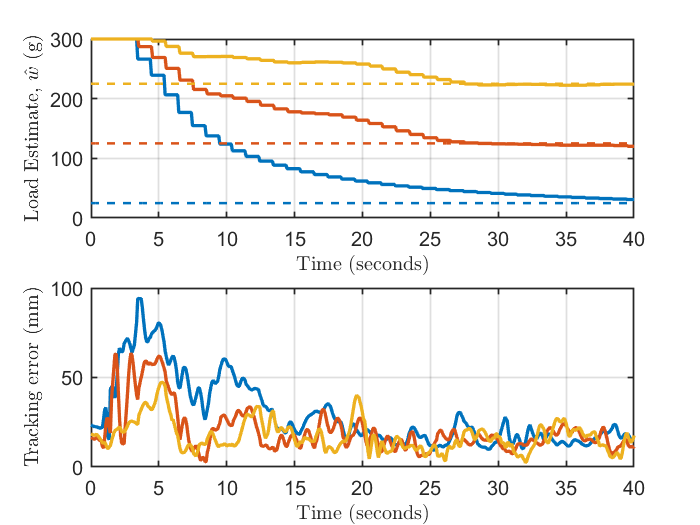}
    \caption{Experiment 3 Results: Periodic trajectory following with an unknown payload. 
    The payload estimate over time (top) and tracking error over time (bottom) are shown for three trials with payloads of 25g, 125g, and 225g.
    Results for the 25g payload trial are shown in blue, results for the 125g payload trial are shown in red, and results for the 225g payload trial are shown in yellow.}
    \label{fig:control-unknown-load}
\end{figure}

To evaluate the efficacy of the combined control and load estimation method summarized by Algorithm~\ref{alg:mpc}, we measured the manipulator's performance in tracking a periodic reference trajectory when the payload is not known.
Once again three trials were conducted with payloads of 25, 125, and 225 grams.
The periodic reference trajectory was a circle with a diameter of 200~mm. 
Note that this trajectory was not part of the training data.
The KL-MPC controller was run at 12~Hz and $\hat{w}$ was updated according to the same parameters as in Experiment~2 (${N_e = 12 , N_w = 30 , N_r = 360 }$). 
% In all three trials, the value of $\hat{w}$ converged to within 25~grams of the actual payload in less than 25~seconds.
% The tracking performance also improved over time, with the tracking error decreasing to less than 30~mm after approximately 20~seconds.
Results of this experiment are shown in Fig.~\ref{fig:control-unknown-load}.

%%%%%%%%%%%%%%%%%%%%%%%%%%%%%%%%%%%%%%%%%%%%%%%%%%%%%%%%%%%%%%%%%%%%%%%%%%%%
% Experiment 4: Automated object sorting
%%%%%%%%%%%%%%%%%%%%%%%%%%%%%%%%%%%%%%%%%%%%%%%%%%%%%%%%%%%%%%%%%%%%%%%%%%%%
\subsection{Experiment 4: Automated Object Sorting (Pick and Place)}
\label{sec:exp4}

\begin{figure}
    \centering
    \includegraphics[width=\linewidth]{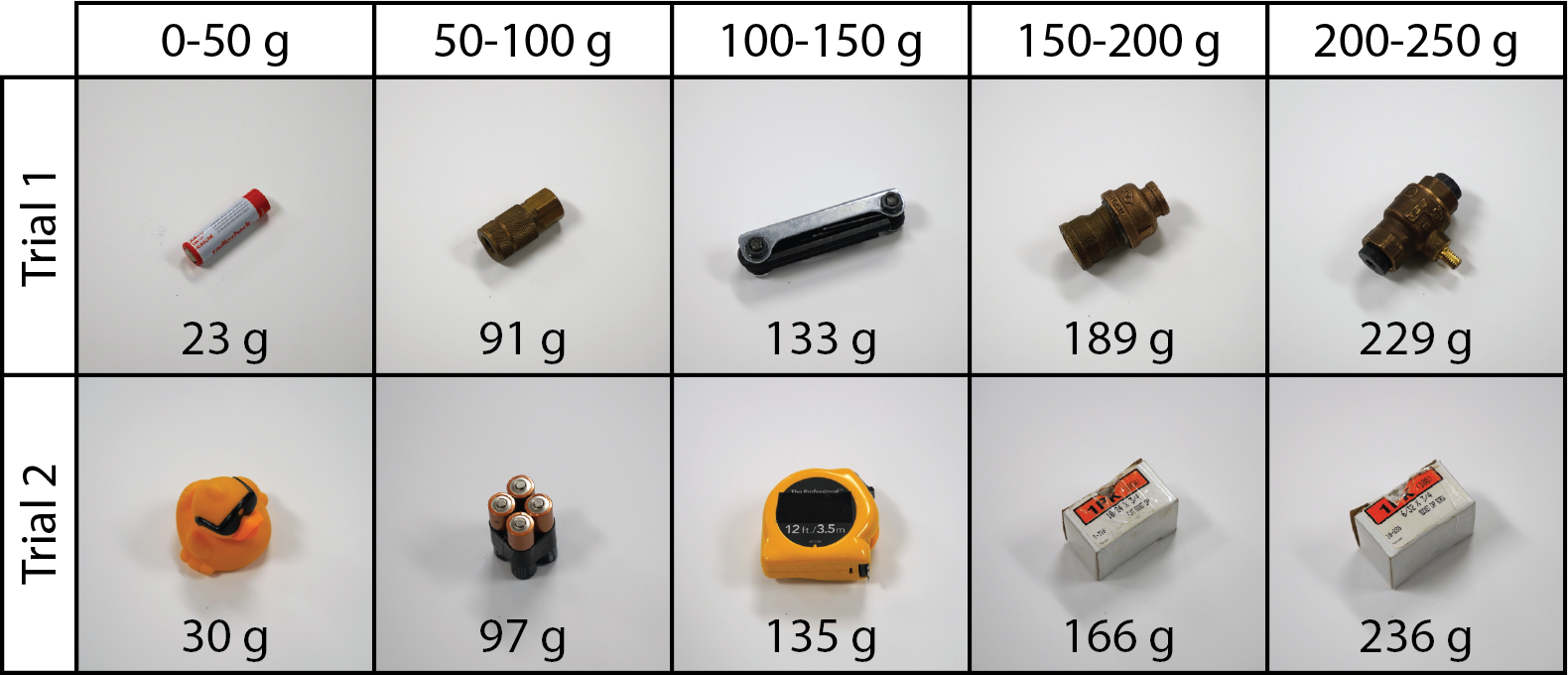}
    \caption{Objects used for Experiment 4: In each trial, the soft manipulator sorted a set of five objects according to their mass, based on an estimate computed by the online observer described in Section~\ref{sec:loadest}. The set of objects used for each trial are separated by row, and the mass of each object is written below it.}
    \label{fig:objects}
\end{figure}

The load estimation algorithm and KL-MPC controller were utilized to perform automated object sorting by mass.
Five objects were selected, each with mass between 0 and 250~grams, and five cups were placed in front of the manipulator, each corresponding to a 50~gram interval between 0 and 250~grams (i.e. 0-50, 50-100, etc.).
The range from 250-300 grams was not used for this experiment, because such loads too severely reduce the workspace of the robot.
The objects used and their masses are shown in Fig.~\ref{fig:objects}.
% Five cups were placed in front of the manipulator, each corresponding to a 50~gram interval between 0 and 250~grams, and five objects were selected, each of a mass that fell into one of the 50 gram intervals.
Given one of these objects, the task was to place the object into the cup corresponding to its mass.
% to the range within which the mass of the object falls.
For each trial, a human assists the manipulator with grabbing the object, then the manipulator performs KL-MPC with load estimation (Algorithm \ref{alg:mpc}) while following a circular reference trajectory for 15 seconds.
After 15 seconds, load estimation stops, and a ``drop-off'' reference trajectory is selected that will move the end effector towards the cup corresponding to the most recent payload estimate.
The manipulator then uses KL-MPC to follow the ``drop-off'' trajectory and deposits the object into the cup.
This cycle repeats until all 5 objects are sorted into the proper cup.
Using this strategy, the manipulator properly sorted 5 out of 5 objects in 2 separate trials, using a different set of objects each time (see Fig.~\ref{fig:objects}).
Footage of these trials can be seen in the supplementary video file.
% \footnote{\href{https://youtu.be/g2yRUoPK40c}{https://youtu.be/g2yRUoPK40c}}.

\section{Discussion and Conclusion}
\label{sec:discussion}

This paper uses a Koopman operator based approach to model and estimate a variable payload of a soft continuum manipulator arm and employs this knowledge to improve control performance.
%% Exp. 1: Control with known load
Our work confirms that incorporating knowledge about the payload into the model improves tracking accuracy and makes the controller more robust to changes in the loading conditions.
In Experiment 1, the KL-MPC controller, which incorporated the payload value, reduced the RMSE tracking error averaged over all payloads by  approximately 33\% compared to the K-MPC controller that did not utilize information about the payload and reduced the standard deviation of the tracking error by about 77\% (see Table~\ref{tab:comparison}).
% \David{You can potentially cut the rest of this part. Or summarize:  ``Using the KL-MPC controller, which incorporated the payload value, reduced the tracking error by about XYZ \% compared to the K-MPC controller that did not utilize information about the payload.}
% In Experiment 1, the L-MPC controller displays the largest tracking error across all loads, likely due to the inaccuracy of its underlying linear state-space model.
% Both of the Koopman-based controllers utilize more accurate models which are capable of describing nonlinear system behavior, yet the performance of KL-MPC was better and more consistent than that of K-MPC across payloads.

%% Exp: 2 Load estimation
To automate the process of identifying payload values, we implemented an observer that was able to automatically estimate unknown payloads within 25~grams in a time of about 15~seconds (see Fig.~\ref{fig:load-estimation}). 
It is notable that this approach was capable of estimating loads other than those presented in the training data set that was used during model-identification.
We did not observe over-fitting to the behavior seen under limited loading conditions which suggests that, despite the fact that the approach is data-driven, the identified Koopman model is able to capture the actual physical effect of various loading conditions.

%% Exp 3: Control with unknown load
By combining the estimation, modeling, and control into a single MPC algorithm (Algorithm~\ref{alg:mpc}), we could demonstrate the effectiveness of our approach to improve control accuracy under unknown loading conditions. 
We first tracked periodic trajectories with an unknown payload.
Since the controller needs some time to establish an accurate estimate of the load, the tracking error gradually decreases  over time as the load estimate becomes more reliable. 
After approximately 15~seconds, the tracking error decreased to less than 30~mm, which was about equal to the error with a known load value.

%% Exp 4: Automated object sorting my mass
As a final demonstration, we implemented successful pick-and-(mass-based)-place object manipulation using the same algorithm. 
Unknown objects were successfully sorted by mass, taking advantage of the fact that the payload estimate was accurate enough to choose the correct container for each object and that the tracking error of the ``drop-off'' trajectory was small enough not to miss the cup.
This required a payload estimate accuracy of less than 50~grams, and a tracking error accuracy of less than 45~mm (the radius of the cups).

% various improvements
While the manipulator exhibited sufficient accuracy to complete this task, several modifications could be made to the robot and controller to improve performance even further.
First, the workspace of the manipulator could be greatly enlarged by replacing some of the current acuators with more powerful ones.
This could be done without significantly increasing size or weight just by increasing the diameter of the PAMs \cite{tondu2012modelling}.
Second, a model and controller could be identified with a shorter sampling time, which would enable the model to account for higher frequency behavior and track more dynamic trajectories.
This could be achieved by making upgrades to our computational hardware and optimizing our code.
Even with these changes, the system's inherent stochasticity would limit tracking accuracy, but these improvements would likely enable much more accurate control.
% \David{One thing to comment about would be the amount of hand-tuning and trail-and error that you had to use.  You approach is very general and people might be curious how easy it transfers to other systems or how much they have to fiddle with the parameters of the algorithm (types and numbers of basis functions, etc...). If that is small, you should proudly report it. (sse below)}
% \David{You present two rather incremental ideas.  Can you maybe think of something that is a bit broader. E.g., extending your approach to include more states (dynamics) or estimate other things? (fluid compressibility as a function of temperature)... I have no idea.. (as you can tell)}

%%%%%% Justify the poor trajectory following accuracy across all trials...
% -Weakness of actuators (i.e. control authority)... could use larger actuators, without adding much weight
% -High frequency vibrations (larger than 12 Hz unmodeled)... could use better hardware and coding (i.e. C++), could have higher frequncy model and execute plans over more than one timestep.

%%%%%%% Concluding remarks
% In summary, this work proposes and demonstrates a modeling and control framework for fully autonomous control of soft continuum manipulators under variable loading conditions.
% While this method is validated on one specific pneumatically-actuated soft manipulator here, the approach is not robot specific.
While, so far, our approach has only been validated on one specific instance of a pneumatically-actuated soft manipulator, it should readily extend to other types and classes of soft robotic systems. 
Beyond specifying the inputs and outputs, no system knowledge was necessary in the implementation, and the tuning of algorithmic parameters such as the type and number of basis functions was minimal.
We thus believe that our work lays a foundation towards enabling the widespread use of automated soft manipulators in real-world applications.

% \David{My suggestion for concluding paragraph:  While, so far, our approach has only been validated on one specific instance of a pneumatically-actuated soft manipulator, it should readily extend to other types and classes of soft robotic systems.  [can we say: ``Beyond the number of inputs and outputs, no system knowledge was necessary in the implementation, and the tuning of algorithmic parameters such as the type and number of basis functions was minimal.''],
% We thus believe that our work has the potential to pave the road towards a real-world application of automated soft manipulators.}

% \section*{Acknowledgments}

\ifCLASSOPTIONcaptionsoff
  \newpage
\fi
\newpage

%% Use plainnat to work nicely with natbib. 
\bibliographystyle{plainnat}
\bibliography{references}

\end{document}